\documentclass[11pt,letterpaper]{article}
\usepackage[T1]{fontenc}
\usepackage[utf8]{inputenc}
\usepackage{lmodern}
\usepackage[margin=1in]{geometry}
\usepackage{amsmath,amssymb,amsthm}
\usepackage{longtable,booktabs,array,calc}
\usepackage{microtype}
\usepackage{xurl}
\usepackage[hidelinks]{hyperref}
\hypersetup{pdfauthor={Yiling Wu},pdftitle={What Bayesian Updating Presupposes: Hume's Representational Conditions and Causal Learning},pdfkeywords={Hume; Bayesian causal learning; experiential grounding; concept acquisition; belief; language models}}
\newcommand{\bibentry}{\par\noindent\hangindent=1.5em\hangafter=1}

\newcommand{\real}[1]{#1}
\title{What Bayesian Updating Presupposes: Hume's Representational Conditions and Causal Learning}
\author{Yiling Wu\\ BridgeM, Inc.}
\date{}
\begin{document}
\maketitle
\begin{abstract}
Bayesian causal models explain how expectations change, but they leave a further question open: how does the learner acquire the causal content whose probability changes? I argue that updating alone leaves three conditions of causal judgment unexplained: experiential grounding, general organization, and belief. Hume links these conditions through a felt customary transition, which arises from discriminating resembling cases and supplies the idea of necessary connexion. His arguments against deriving necessity directly from a single interaction or from an act of will support this account of the source. Contemporary causal-theory learning uses perceptual and agency cues to acquire an intervention role. The resulting disagreement is about what that role explains about experienced causal necessity, not about whether the model learns anything. Three elementary constructions distinguish acquisition from assisted execution, lost distinctions from failed access, and posterior agreement from usable deployment. Sampling theories and stable-belief accounts supply further components beyond updating. For models that learn about external events solely through a text corpus and written reports, the evidence is report-mediated rather than firsthand; whether their own transitions are felt is a separate, open question, and the update rule settles neither matter. The paper identifies what a complete explanation must connect: the acquisition of causal content, its availability for general use, and its role in belief. Correct posterior prediction alone establishes none of these connections.

\end{abstract}

\section{The causal judgment that updating leaves to explain}

A learner can become more confident that one event causes another without that change in confidence explaining how she came to understand causation. This is the paper's central distinction. Bayesian updating explains how probability changes over a domain. Hume's account also asks how experience supplies causal content, how particular occurrences are organized under relevant similarities, and what makes an expectation a belief. An explanation that keeps the change in expectation but omits these conditions explains less than the formation of a causal judgment.

The claim has identifiable interlocutors. Cheng (1997, 397--398) explicitly presents causal-power theory in relation to Hume and Kant. Her account derives particular powers from covariation, given general assumptions about production and prevention, and Lu et al. (2008, 955--957) incorporate its generating functions into Bayesian causal models. Griffiths and Tenenbaum (2009, 662--663, 708--709) place covariation theories in a Humean tradition and explain induction through Bayesian inference guided by causal theories, including theories that can themselves be learned. These are substantial explanations of causal learning. What needs settling is which parts of Hume's problem they retain, which they replace, and which they leave unresolved.

Formal epistemology engages Hume differently. Howson's Hume--Ramsey project develops an inductive logic conditional on initial plausibility assumptions (2000, 2--4, 239--240). Earman distinguishes descriptive from normative Bayesianism and examines whether personal-probability induction is compatible with Hume's concerns (1992, 56, 94--97). Neither project should be made to promise a developmental psychology. Their relevance lies in a further inference: a successful account of inductive warrant does not by itself explain how the content to which its norms apply is acquired. The comparison is about conceptual connections among explanatory projects, not a historical claim that Bayes formalized Hume's causal psychology.\footnote{Stigler (2013) reconstructs an anti-Humean context for Bayes's essay. Price's later use of the result against Hume is documented; Bayes's own motivation is the more speculative link. The conceptual comparison here neither denies that history nor depends on establishing it.}

I defend a specific insufficiency thesis. \textbf{Conditionalization over an interpreted domain does not explain the experiential grounding, general organization, or belief role of the representations to which it is applied.} When an account claims to explain the formation of a causal judgment, these become substantive explanatory requirements. A mathematical probability space need not contain Humean impressions; but interpreting its evolution as the development of the kind of judgment Hume describes requires an account of the three conditions. A theory may supply that account or defend an alternative conception of causal content, but it cannot obtain either result from correct posterior predictions alone.

The strongest contemporary models sharpen the issue. Hierarchical learning acquires inductive preferences; sampling and exemplar processes explain selective consideration; Leitgeb relates categorical belief to stable high credence; and Goodman, Ullman, and Tenenbaum (2011; hereafter GUT) even give an initially uninterpreted predicate a learned intervention role. Each of these is an addition to an updating operation: it specifies perceptual resources, representational languages, search mechanisms, semantic roles, or doxastic constraints. Together they support the need for a fuller explanation. What remains in dispute is why learning an intervention role should explain the learner's sense of causal necessity. Hume argues that neither a single observed interaction nor consciousness of willing discloses a power binding cause to effect, and GUT's perceptual and agency starting points reopen exactly these questions. Section 4 weighs the competing reasons, including evidence for perceptual causality, instead of imposing a historical theory as a condition on every modern model.

The paper's contribution is to connect this disagreement about content with two further questions about evidence. First, interpreting a mature state is not yet explaining how it was acquired. Second, a demonstrated posterior or a successful response does not establish every use or commitment attributed to that state. The three conditions must be connected along one and the same learning history. This yields a definite verdict on numerical-program learning: it can explain the acquired possession of usable resources within a fixed language, but it neither settles the origin of primitive understanding nor establishes belief in the propositions those resources express.

Sections 2 and 3 state the Humean conditions and distinguish their work from constraints on priors. Sections 4--6 take up experiential grounding, general organization, and belief in turn. The elementary propositions appear alongside the conditions they diagnose; they mark limited evidential boundaries, whereas the impression-source thesis is a philosophical claim defended from Hume's text. Section 7 applies the distinctions to language models, and section 8 explains why the three conditions must concern the same acquired judgment. Technical details on usable information and the finite stability account are given in the appendices.

\section{Three connected conditions in Hume}

Hume's account combines three claims: causal transition is produced by custom, not by an argument establishing its rational necessity; experienced conjunctions shape the strength of expectation; and belief is a manner of conceiving a content, not an additional idea. On the first point he writes that the transition is ``not determin'd by reason, but by certain principles, which associate together the ideas of these objects'' (T 1.3.6.12, SBN 92).\footnote{References to the \emph{Treatise} (T) give book, part, section, and paragraph numbers in Norton and Norton's critical edition (2007); references to the \emph{Enquiry concerning Human Understanding} (EHU) follow Beauchamp's critical edition (2000). SBN gives page numbers in the Selby-Bigge editions revised by Nidditch (1978 and 1975). A note is cited by ``n'' and its number (e.g., EHU 7.15n13).} Uniform experiences strengthen the customary tendency and contrary ones divide it (T 1.3.12.2--8, SBN 130--134). Belief gives a conception a distinctive force and influence (T 1.3.7.7, SBN 628--629). Taken together, these claims say more than the advice to let experience guide expectation.\footnote{Hume also calls causal inferences reasonings; his explanatory claim is about what produces their transition. A normative Bayesian rule need not be an algorithm that a reasoner consciously executes, so denying reason a psychological role does not by itself refute the rule. The correspondence defended here holds at the level of evidence-sensitive expectation; it does not claim that the two psychological explanations are the same.}

\textbf{Experiential grounding} concerns the historical acquisition of causal content. The claim that carries the weight is local: Hume locates the source of the idea of necessary connexion in the felt determination that accompanies customary transition. In the \emph{Enquiry}, ``this customary transition of the imagination'' is ``the sentiment or impression, from which we form the idea of power or necessary connexion'' (EHU 7.28, SBN 75). The claim is about what supplies the idea, not merely about some event that happens to cause its acquisition. Hume's inquiry into the impression thereby determines what positive content his account permits.

\textbf{General organization} concerns the respects in which occurrences are compared and the availability of relevant particulars. Other ideas can be present ``only in power,'' available as a ``present design or necessity'' calls for them (T 1.1.7.7, SBN 20--21). An alternative triangle can correct an inference drawn from the exemplar currently entertained, and the same image can function under different general terms (T 1.1.7.8--9, SBN 21--22). The white marble globe further illustrates comparison across shape and color: general use requires attention to a respect of resemblance, not merely the activation of an undifferentiated similar item (T 1.1.7.17--18, SBN 24--25). In the first definition of cause, the scope of ``all the objects resembling'' therefore requires both a range of instances and a relevant respect (T 1.3.14.31, SBN 170).

This is a modest dispositional reading of T 1.1.7. Garrett's revival sets develop the account (1997, 24, 63--64), but naming a set does not explain how its appropriate members or dimensions become available. Broughton distinguishes discrimination, feature recognition, and general signification while reconstructing the circularity challenge to prior recognition of resemblance (2000, 282--289). Hume himself acknowledges the difficulty of explaining how custom operates (T 1.1.7.11, .15, SBN 22--24). A contemporary process model can therefore improve on the mechanism without making the explanatory requirement disappear.

\textbf{Belief} concerns the manner in which a content is entertained. Firmness, steadiness, feeling, and influence distinguish believing from merely supposing. A transition can occur within hypothetical reasoning without belief in a real existence (T 1.3.4.2, SBN 83), and T 1.3.8.11 explicitly describes an associative transition with ``no belief nor perswasion'' (SBN 103). Understanding and using a causal claim inside a supposition therefore need not amount to endorsing it, and the distinction requires no new constituent in the claim's content.

The source passage connects elementary discrimination with an enlivening determination. Hume writes that after observing resemblance in enough instances ``we immediately feel a determination of the mind to pass from one object to its usual attendant, and to conceive it in a stronger light upon account of that relation'' (T 1.3.14.20, SBN 164--165). Observed resemblance and the collection of ideas give a minimal organization of cases; the resulting determination is an internal impression; and the transition strengthens the conception. This initial organization requires the discrimination of similar instances, not the full general condition described above.

Animals can modify customary expectations through experience without first deriving a universal conclusion (T 1.3.16.7--8, SBN 178). In endorsed causal attribution, the fuller condition concerns the range and respect of comparison articulated by general ideas, together with the universal commitment of the first definition. The source passage connects the elementary discrimination and the enlivening transition with this later use; it does not make a fully articulated causal judgment a prerequisite for acquiring the idea of necessity. The originating belief in the attendant can likewise be distinguished from the endorsement of a causal attribution. Later use requires no fresh impression of the original kind: Hume allows conviction in established principles to remain even after the impressions that gave rise to them are forgotten (T 1.3.4.3, SBN 84).

Two qualifications remain on this local reading. First, the conditions do not guarantee that every causal attribution is true: a generalization may be mistaken, or an association prejudicial. Second, the account does not settle every metaphysical question about mental feeling. It does, however, take the felt character of the source seriously: a bare transition, without an account of its status as an impression, is not automatically the source Hume describes. Wu (2026) defends this interpretation of the two definitions in detail. Buckner (2024) sets empiricist faculty psychology in a broader exchange with contemporary learning theory; the comparison here isolates the acquisition and use of causal content.

Hume's theory also contains normative evaluation. He introduces ``general rules, by which we ought to regulate our judgment'' and distinguishes corrective from prejudicial generalization (T 1.3.13.11--12, SBN 149--150; see also T 1.3.15, SBN 173--176). The relevant contrast with Bayesian epistemology is therefore between particular explanatory tasks, not between a wholly descriptive Hume and a wholly normative Bayesianism.

\section{Priors constrain credence; sources constrain content}

For a finite hypothesis space, Bayes's theorem states

\[P(h\mid e)=\frac{P(e\mid h)\,P(h)}{\sum_{h'}P(e\mid h')\,P(h')}.\]

This identity relates probabilities on a specified domain. A normative theory explains when an agent ought to set later credences equal to earlier conditional probabilities, or adopt some other evidence-sensitive revision rule. A cognitive theory also interprets the hypotheses and specifies which aspects of the learner's organization the model explains. A decision model adds available actions, losses, and a response rule. Jones and Love (2011, 183--186) distinguish these levels and call for explanations of the representations and learning processes that Bayesian inference assumes; their criticism already covers the construction of hypothesis spaces. The additional question here is narrower: whether a learned causal role also supplies the experiential source of necessity content, and how that acquired content enters general use and belief.

Howson's and Earman's discussions of Hume belong mainly to the justificatory project introduced in section 1. A proposed norm can succeed without explaining how its user acquired the concepts to which it applies. The converse also holds: identifying the causal history of a confidence assignment does not establish that it is rational. Joyce's (1998) accuracy-based argument, for example, addresses the justification of probabilistic coherence. It is to be judged by its normative premises and consequences, not by whether it describes a child's acquisition of a hypothesis language.

The objection that Bayesian priors are arbitrary therefore misses important targets. Williamson (2010, chs. 1--4, 9) advances principles of Probability, Calibration, and Equivocation, which tie admissible belief to evidence and limit unsupported commitment. He also addresses the language relative to which probabilities are assigned. Landes and Williamson (2013, 3529, 3565--3566) reconstruct this approach explicitly: evidence constrains the admissible probability functions, and equivocation further regulates the choice among them. Revised assignments must satisfy the applicable norms, and conditionalization is not the universal update rule. This is a substantive account of evidential constraint.

Hume's question about origins concerns a different relation. Calibration asks which probabilities accord with the evidence for claims that are already interpreted. The source question asks what makes the learner's representation a representation of the attributed feature or causal relation. Evidence can constrain both relations, and a larger theory can address both. Still, an answer about admissible numerical assignments does not by itself settle the content of the propositions that receive them. Recognizing this distinction neither denies Williamson's treatment of language nor demands that a normative theory become a developmental psychology.

Hierarchical Bayesian learning addresses a further objection: inductive preferences need not remain fixed. Kemp, Perfors, and Tenenbaum (2007, 310--312, 317--319) show how observations across categories can teach expectations about feature variability, which then guide the learning of a new category. Their higher-level variables support priors for lower-level learning, so what is prior for one task can reflect earlier evidence. They distinguish the model's possible hypotheses from those it currently entertains, and they do not present their numerical implementation as a psychological processing algorithm (309--312). These are genuine learning achievements, not merely an arbitrary prior hidden under another name.

The remaining question has to be located precisely. In that model, objects, categories, and features already have a specified interpretation, and learning changes expectations about their structure. If the target is the acquisition of an overhypothesis about those features, the model addresses it directly. If the target is how a learner first comes to represent the features themselves, the explanation needs further components. Higher levels can explain further structure, but each claimed achievement must say what was initially available and what experience changes. This is a question about the scope of the learning explanation, not a prohibition on learned priors or representational learning.

Hume's discussion of chances makes the role of a specified domain especially clear. He treats ``perfect indifference'' as a condition of elementary chance and accepts the calculus of ``a superior number of equal chances,'' while asking how numerical superiority produces assent (T 1.3.11.4--8, SBN 125--127). The die's causal organization supplies a delimited set of possible outcomes. In his example of a six-sided die, four faces carry one marking and two carry another, so equal chances for the individual faces yield unequal expectations for the markings (T 1.3.11.6, .12--13, SBN 125--126, 128--130). Equal weighting over faces and unequal weighting over grouped outcomes are thus compatible. Hume's ``native situation of indifference'' characterizes an imaginative state; it does not impose equal priors on every description of a problem.

The passage supports a limited comparison. Both the assignment of chances and the strength of the resulting expectation depend on how alternatives are individuated and grouped. Hume's psychological account does not supply Williamson's normative justification for a probability assignment, and Williamson's normative constraints do not claim to reproduce Hume's imaginative mechanism. The problem they share is how evidence bears on an articulated domain; how that domain is interpreted and acquired remains a question that a cognitive explanation may need to address.

\section{Experiential grounding: acquiring the content that is updated}

\subsection{Evidence and the acquisition of an idea}

An experience can contribute to judgment in two ways: it can supply evidence for a claim already understood, and it can supply the material from which an idea is first formed. Hume's general copy principle makes the second role explicit for simple ideas (T 1.1.1.7, SBN 4), and his example of a person blind from birth shows its intended force (EHU 2.7, SBN 20). These general examples clarify the distinction, but the argument here requires only the local application to causal necessity established in section 2.

Jeffrey's candlelight example separates the two roles especially well. A person looking at a piece of cloth by candlelight revises the probabilities that it is green, blue, or violet, although the experience need not yield any newly certain proposition about how it looks (1990, §11.1, 165--166). Jeffrey conditioning accommodates this kind of evidential impact. The color alternatives already have a place in the person's assessment: given revised probabilities for them, Jeffrey's rule propagates the change to other propositions while holding the relevant conditional probabilities fixed. It does not explain how the person first acquired the color content of those alternatives. Allowing a nonpropositional experience as input therefore does not by itself answer the source question. The objection is not that Bayesian evidence must always arrive as a verbal proposition. Jeffrey also treats experience-acquired confirmational relations as part of word mastery (§11.10, 179--180). That broader account deserves recognition, but it still does not identify those relations with the felt source of Hume's idea of necessity.

In the causal case, the distinction falls within the mental transition itself. Repeated resemblance leads the imagination toward the usual attendant, and a probability model can represent a corresponding change in prediction. But Hume also identifies an impression of that customary determination as the source of the idea of necessary connexion (EHU 7.28). A description of the transition's input and output can capture the evidential pattern while leaving out this content-producing role. The omission matters when the explanation on offer is of how a thinker acquires causal understanding, and not only of how an understood causal hypothesis gains support.

Copying offers a route outside hypothesis testing: one need not first formulate and confirm a hypothesis using the very simple idea whose acquisition is at issue. Fodor's paradox targets acquisition through such prior formulation and testing, and Fodor himself allows experience-sensitive, noninductive concept acquisition (2008, 132, 137--140, 144--145). Hume's proposal thus identifies a kind of acquisition that an updating explanation needs to supplement; it does not refute Fodor simply by calling that process learning. Whether copying gives an adequate account of the resulting content remains a substantive question. What matters here is that conditioning on hypotheses cannot by itself explain this noninferential contribution to understanding them.

\subsection{Causal powers, learned theories, and the strongest counterexample}

Cheng states her alternative starting point explicitly: ``I assume that the power PC theory describes an innate component of the process of causal induction'' (1997, 397). General knowledge of production and prevention licenses inference to a particular power; it is not acquired by copying an impression of one's customary transition (397--398). Lu et al. (2008, 957) explicitly identify the Humean and Kantian strands when they incorporate Cheng's causal generating functions into their Bayesian account. The empirical estimation project can succeed from that starting point. Its relation to Hume, however, involves replacing the account of the content's source, not simply giving a formal rendering of the same explanation.

Griffiths and Tenenbaum extend learning to the background causal theories themselves. They end the explanatory regress in basic assumptions about causality, and they distinguish computational learning from possible psychological implementations, including Monte Carlo and particle methods (2009, 708--709). Their account is not limited to fixed theories, and it does not omit selective processing. The question is where their learned structure begins, and what gives their initial causal resources the interpretation attributed to them.

GUT presents the strongest challenge to the claim that causal content must simply be supplied. Its theory-learning model starts with logical resources, probabilistic dependence, perceptual input analyzers, and a latent predicate A. GUT's Table 1 describes A as initially meaningless; learned laws give it an intervention role. The perceptual supports can include agency, launching, and self-efficacy cues. This is genuine learning of a conceptual role, not merely the reweighting of an intervention predicate that is already interpreted: a fixed language can support a change in the significance of one of its expressions.\footnote{References to GUT follow the authors' accepted manuscript. They cite its Table 1 and the sections ``Theories of causality,'' ``Theory induction,'' and ``Discussion,'' because the manuscript's pagination differs from that of the printed article.}

These perceptual starting points bring GUT into contact with Hume's argument, and not merely with an extra requirement he happens to impose. In EHU 7.6--8 (SBN 63--64), the motion that follows a billiard-ball impact presents a succession without revealing a power that makes the second motion necessary. Hume's reason is that grasping that power in a single occurrence would allow its effect to be anticipated independently of prior experience. In EHU 7.9--20 (SBN 64--69), he tests the alternative appeal to consciousness of willing. Our sense of command does not independently disclose its limits or the bodily intermediaries through which action proceeds. His paralysis example separates apparent command from successful movement: the efficacy of willing has to be learned from its results. The argument targets the supposed disclosure of necessary power, not the existence of perceptual or agency cues.

GUT can reply that its learners need only fallible cues from which an intervention role is learned, not prior insight into an infallible productive tie. Hume himself allows the feeling of effort to contribute to ordinary, imprecise ideas of power, while denying that it yields the precise idea sought in the inquiry (EHU 7.15n13, SBN 67; 7.29n17, SBN 77--78). His negative argument therefore does not exclude every contribution from agency or bodily experience.

The Michottean tradition of research on perceptual causality supplies further serious counterpressure. Leslie and Keeble (1987, 265, 282) found that infants responded differently to role reversals in launching events than in matched delayed events. Kominsky and Carey (2024, §5) distinguish early representations of launching and of triggering, with different evidence for the encoding of causal roles. These findings support early event organization, but they establish neither infants' phenomenology nor an insight into necessary power. A Humean account must accommodate this sensitivity. The question that remains is whether this perceptual organization, or the intervention role learned from it, explains the necessity attributed to the connection.

There is a positive reason to keep this question open. Identifying which intervention changes an outcome and experiencing a connection as compelling are different things to explain. Visual-adaptation experiments with adults support a perceptual contribution to causal appearance, since the effect depends on retinal location (Rolfs, Dambacher, and Cavanagh 2013). Perceived production puts pressure on the stronger denial, in EHU 7.6, that the single succession yields any relevant impression. A theory of causal understanding thus has a psychological phenomenon to explain, beyond fitting predictions. Hume's candidate explanation locates the positive content of necessity in the determination felt after repeated resembling cases (EHU 7.28). It offers an account of how experience can alter the apparent force of a connection without uncovering a new intrinsic quality in either event. The felt transition is the proposed source of the idea of necessity, not a perception that guarantees either an intrinsic necessitating tie or the truth of the attribution. GUT learns an intervention role, but it does not model the origin or the experienced character of its proposed perceptual inputs. Whether perceptual organization or felt customary determination gives the better account of the source, and whether that experience constitutes causal content, remain substantive disputes. Correct posterior predictions cannot decide them, because the rival accounts can agree on those predictions.

A conceptual-role theorist can reject the further Humean requirement. Piantadosi (2021) models meanings as emerging through relations within a learned conceptual system, without requiring every primitive to arrive with its eventual semantics; he leaves the formal integration with perception and action to further work (§6.1). My reply is not that uninterpreted symbols can never acquire content, but that conceptual-role semantics is an additional account of what makes a state meaningful. If it replaces felt experiential origin, the theory has changed its account of causal content; if it supplements that origin, the acquisition history still needs explaining. Conditionalization alone decides neither issue.

The criticism is therefore aimed at explanatory sufficiency. An account of useful intervention concepts can succeed on its own terms. To extend it to a full psychological explanation of human causal understanding, one must explain how causal appearance, learned intervention roles, and endorsed judgment are related. Hume's source inquiry offers one account of their connection. A rival may explain the experienced aspect separately instead of making it constitutive of causal content, but an updating calculation alone supplies neither explanation.

\subsection{What an acquisition demonstration must establish}

Four distinctions help to locate the claim without multiplying the paper's conditions. The modeler's description domain is not the learner's interpreted scheme; an expression within that scheme is not necessarily a candidate the learner actually generates; and a generated candidate need not be available for a target operation. The first contrast bears on experiential grounding, the others on general organization. These are distinctions within an explanation, not four psychological faculties.

There is also a temporal requirement. A correct interpretation of a mature state does not explain how the state came to have that role. Color and ingredient, for example, can covary in every observed treatment, and a final classifier that fits those observations can track either feature. Further cases that separate them may settle the attribution, and so may an independently established learning history. A mature interpretation in terms of ingredients cannot simply be read back into every earlier stage. Information, its usable role, and its history must jointly support the claimed content (Shea 2018; Harding 2023). This is one way to make the grounding question operational without confusing it with a complete metaphysical theory of meaning (Harnad 1990).

The elementary constructions below identify what certain demonstrations leave open. They concern usable operations and discriminations; Hume's impression-source thesis does not follow from their mathematics.

Consider the task \(c_\theta(x)=x\oplus\theta\), where \(x,\theta\in\{0,1\}\) and \(\oplus\) is exclusive disjunction. The hidden parameter is drawn independently and uniformly in each episode. One noiseless example \((x,y)\) identifies \(\theta=x\oplus y\). Two systems, C and D, share a posterior register \(p\), initially \(1/2\), which ordinary conditioning updates to \(\theta\); both answer probability queries by returning \(p\). They also share a classifier \(d(m,x)=x\oplus m\) and a memory bit, initially \(m=0\). After the example, C writes \(x\oplus y\) into \(m\), while D leaves \(m\) unchanged. A supplied-state test instead writes \(m=\theta\) externally, bypassing and disabling both learning procedures during testing.

\textbf{Proposition 1. Supplied-state success does not identify acquisition ability.} Both systems classify perfectly with the supplied state. With the example alone, C is perfect, whereas D has expected accuracy one half, although their posterior trajectories and probability reports are identical.

\emph{Proof.} The supplied write gives both classifiers \(m=\theta\). Under ordinary acquisition C computes that value, while D retains zero and succeeds exactly in the episodes with \(\theta=0\). Both probability registers receive the same update. \(\blacksquare\)

The distinction concerns the acquisition of the operative setting, not of a new semantic primitive; both systems acquire information in \(p\). It is a minimal instance of the difference between a skill and the capacity to acquire it (Chollet 2019). Its application to a supplied parser, rule, or interpreted input is conditional: if an explanation claims to show that such a resource was acquired, testing downstream success after supplying the resource bypasses the very transition to be explained. GUT's supplied input analyzers do not invalidate its learning of the intervention role; they delimit which acquisition its result demonstrates.

Whether a capacity is identified depends on the admissible models and observations. Let \(\mathcal{M}\) be the model class, \(T_B(M)\) the distribution of complete observable transcripts under condition \(B\), and \(K(M)\) the target capacity. Identification requires

\[T_B(M)=T_B(M')\ \Rightarrow\ K(M)=K(M')\qquad (M,M'\in \mathcal{M}).\]

The supplied-state condition fails this test for a class that admits both learning procedures. Evidence about architecture, intermediate observations, or an acquisition condition can distinguish them. There is therefore no general impossibility in identifying construction, as opposed to selection, from behavior. The evidential boundary is relative to what the test observes and which alternatives remain open.

A stronger limitation arises when the complete accessible state loses a required distinction. Let \(r:H\to S\) map histories to all the resources available when answering, let \(Q\) be the queries, and let \(\tau:H\times Q\to A\) give the single correct answer. No new input may distinguish the histories during answering. Any random seed correlated with the history is counted as part of the state, and residual randomness has the same conditional distribution given the state and the query.

\textbf{Proposition 2. A collapsed task distinction prevents uniformly correct answering.} If \(r(h)=r(h')\) but \(\tau(h,q)\neq\tau(h',q)\), no deterministic procedure restricted to that state and query succeeds in both cases. A randomized procedure has average zero-one error at least one half on their equal mixture.

\emph{Proof.} The inputs are identical, so a deterministic procedure returns the same answer in both cases. A randomized procedure gives the same answer distribution, and the probabilities it assigns to the two distinct correct answers sum to at most one. \(\blacksquare\)

If later success depends on experience supplying the missing discrimination, the explanation must account for how it arrives. The result does not require every new task to introduce a semantic primitive. Nor does it turn a failed response into evidence that information was destroyed: an unsuccessful procedure may ignore a distinction that is still present in its accessible resources.

The event algebra makes the narrower limit of conditioning explicit. For a probability measure \(P\) on a fixed event algebra \(\mathcal{F}\), and \(E\in \mathcal{F}\) with \(P(E)>0\),

\[P_E(A)=\frac{P(A\cap E)}{P(E)},\qquad A\in \mathcal{F}.\]

This operation changes a measure on the same algebra. It neither introduces an event outside the algebra nor supplies an interpretation of its elements. A larger architecture may construct new expressions, acquire conceptual roles, or extend the domain, and open-minded Bayesian frameworks explicitly accommodate new hypotheses (Wenmackers and Romeijn 2016). A mathematically fixed space can also contain dispositions whose availability changes through learning. What can be defended, therefore, is that the updating operation is insufficient for acquisition, not that Bayesian theories of acquired representation are ruled out. For the local Humean target, such a theory must also explain the impression-source of causal necessity or defend its replacement.

\section{General organization: similarity, retrieval, and usable resources}

\subsection{Bayesian counterparts to general employment}

A Bayesian hypothesis can be general: the claim that Newtonian mechanics is true concerns indefinitely many physical systems without enumerating them. Hume's abstract ideas likewise need not present every instance at once. The relevant question is how a bounded thinker brings a suitable case or alternative into use, and how it is compared in the appropriate respect.

Shi et al. (2010, 447--449) provide an especially close counterpart. Under their importance-sampling correspondence, stored exemplars sample the prior and similarity weights are proportional to likelihoods, so particular stored cases can support approximate Bayesian inference. This comes closer to Hume's dispositional availability of particulars than a static list of hypothesis labels does. It is not the same as Garrett's revival-set interpretation, since the computational proposal specifies one way in which stored samples contribute to an inference.

What remains is the question of the respects of comparison. In a given implementation, stimulus representations and weighting rules determine which exemplars matter. Shi et al. explicitly allow similarity functions to vary with the task, and they discuss dynamic hypothesis spaces through particle filtering (448, 461--462); a criticism claiming that their whole theory fixes every feature forever would overlook these provisions. The Humean question is whether the model explains the acquisition and use of the relevant dimension in the case at hand. Learned attention or a learned metric can answer part of it, and such mechanisms would amount to an explanation of general organization beyond the conditionalization rule.

Accounts of hypothesis search add another kind of counterpart. Dasgupta, Schulz, and Gershman (2017, 3--5) model bounded Metropolis--Hastings search with a proposal process, evaluation, and path-dependent traversal; the evidence stays fixed during the modeled search. Sanborn and Chater (2016, Table 1 and Box 1) explain how limited samples can support judgments without an explicitly represented probability table. Both supply mechanisms for selective consideration. The developmental question that remains is how a learner acquired the proposal resources and their task-relevant interpretation, which is not the same as denying that the proposed search generates hypotheses.

The comparison is positive but demanding. Bayesian process accounts can explain aspects of general organization more explicitly than Hume does. This supports the insufficiency thesis about updating, because the explanation comes from additional representations, sampling operations, and selection rules. To explain a single acquired judgment, the resources whose learning has been modeled must be the resources that those operations actually recruit.

\subsection{A posterior is not an additional capacity}

The binary task of section 4 gives a minimal demonstration. Let systems A and B share the same prior, evidence encoding, posterior update, and register \(p=P(\theta=1\mid e)\), and let both return \(p\) to probability queries. After an informative example, A classifies by \(D_A(p,x)=x\oplus p\), while B uses \(D_B(p,x)=x\). The target and the scoring rule are the same.

\textbf{Proposition 3. Equal posterior trajectories do not determine deployment for an additional operation.} A and B give identical posterior reports after every common evidence sequence on which conditioning is defined. A classifies perfectly; B fails whenever \(\theta=1\).

\emph{Proof.} The initial registers and their update functions are identical, as are the reporting functions. After informative evidence \(p=\theta\), so A returns \(c_\theta(x)\) while B returns \(c_0(x)\). These classifiers disagree on both inputs when \(\theta=1\). \(\blacksquare\)

The target here is the inference from probability reports to a further capacity. A full Bayesian decision model that specifies actions, losses, and a decision rule already explains the behavior under those joint assumptions, and the construction does not refute such a model. It shows why the model's operative route is an additional premise, not something established by matching posterior reports. B may be an implausible model of a competent reasoner, but excluding it requires exactly the use constraint whose presence was in question.

Unrestricted recoverability, recoverability within a predictor family, and actual deployment are distinct. Xu et al.'s (2020) predictive \(\mathcal{V}\)-information formalizes the middle notion. In the example, both systems preserve the same bit, and a suitable common predictor family can recover it equally well, yet their executed classifiers differ; Appendix A gives the definitions and the calculation. This is why the collapse result of Proposition 2 is stronger than a failed controller: a collision of complete accessible states defeats every decoder, whereas a failure of deployment need not destroy information.

Probing meets the same evidential limit. Control tasks reveal how much a probe itself can learn (Hewitt and Liang 2019), and extraction and causal use call for different evidence (Belinkov 2022). Successful decoding also needs justification before it establishes represented content (Ritchie, Kaplan, and Klein 2019). For present purposes, a useful intervention must connect the recovered distinction to the learner's own operation. Its success then supports general organization, but it does not independently establish the experiential source or the endorsing role of that distinction.

\subsection{Which acquisition a fixed language can explain}

A fixed interpreted language can support the acquired possession of usable resources. Membership in its full compositional closure does not show that a resource has been constructed, selected, or made available. Perfors (2012) distinguishes latent from explicit hypothesis spaces, and Margolis and Laurence (2011) explain how evidence can guide transformations before a complete target representation is first entertained. These are genuine gains in general organization, but they are not automatically gains in the primitive understanding that Fodor distinguishes from successful application.

The dispute over numerical learning illustrates the point. Piantadosi, Tenenbaum, and Goodman (2012, 205--207, 211--214) specify programs, evidence-sensitive evaluation, search, and a numerical response procedure. Rips, Asmuth, and Bloomfield (2013, 321--330) grant computational feasibility but dispute the claimed conceptual discontinuity and the claimed psychological interpretation. Carey (2009, 474--477, 511--513) describes Quinian bootstrapping through partially interpreted placeholders, analogy, and modeling. The question is which change the computational result demonstrates.

My verdict turns on the criterion of possession. If possession means inclusion in an unchanged interpreted closure, generating an element of that closure cannot establish the discontinuity at issue. Under the criterion of available use, by contrast, the model does explain acquisition, because its learning process changes what the modeled learner can recruit. This identifies the standard under which the discontinuity objection succeeds; it does not attribute to Rips and colleagues a general theory of closure. \textbf{Possession of usable resources} is a criterion for the acquisition of a capacity, and it deliberately does not equate that target with resolving Fodor's paradox about understanding.

Rule and Piantadosi's (2025, 31--33, 49--59) resource-bounded proposal is a close antecedent. Chunks can increase practical expressivity, and posterior changes can lower the cost of selecting and representing a concept, thereby changing the degree to which it is possessed. Their proposal already connects belief fixation, access, and acquisition, so I cannot claim those connections as new. The further question is what follows when that gain is used to explain a causal judgment. A lower selection cost does not by itself establish the experiential source of the selected content or its role as categorical belief.

The additional requirements are therefore concrete. Proposition 1 asks whether the learner established the usable resource or received it through the test. Proposition 3 asks whether posterior information governs the further operation attributed to it. The next section distinguishes using a concept from believing a proposition that contains it: a thinker may competently use an acquired concept within a hypothesis she rejects. This does not deny that posterior changes can help a learner acquire a resource; it separates that achievement from a claim about what the thinker takes to be true.

\section{Belief: what stability explains and what it presupposes}

A calculation made inside a supposition can be accurate, understood, and causally effective without expressing the calculator's belief. A scientist can predict from a model she regards as false; a system can report another agent's confidence or follow a conditional fiction. In Hume's terms, transition and conception do not by themselves supply the manner of belief (T 1.3.4.2; T 1.3.8.11). The further question is what gives the content authority in ordinary expectation and action.

A genuine credence is already a doxastic attitude. If a probability has been independently identified as the agent's own credence, that attribution answers part of the question, but a numerical register or a correct report does not establish it unaided. Two procedures can likewise agree on every probability inside a hypothetical model while one carries its conclusions into ordinary planning and the other confines them to simulation. A difference in the manner of employment need not add a new constituent to the proposition represented.

Leitgeb's Humean thesis gives a substantive account of the relationship between graded and categorical attitudes (2015; 2017, ch. 2). In a finite formulation, let \(W\) be the world set, \(P\) a probability measure, and \(\mathrm{Bel}\) a consistent belief set. For a threshold \(1/2\le r<1\), the condition is

\[\mathrm{Bel}(A)\ \Leftrightarrow\ P(A\mid E)>r\ \text{ for every } E \text{ with } P(E)>0 \text{ and } \neg\mathrm{Bel}(W\setminus  E).\]

The admitted suppositions must be compatible with the \textbf{whole belief set}; they are not every positive-probability event that intersects \(A\). The condition relates belief to credence. It neither identifies categorical belief with a single high marginal probability nor derives the entire belief set uniquely from that number. Appendix B gives the stable-core formulation and an example of the difference.

The application to Hume is substantial. Stability captures one aspect of firmness and steadiness: support for a believed proposition stays high under the admitted suppositions, not merely at one point. Leitgeb also connects belief with decision-making under specified assumptions about utilities, actions, and context (2017, §§2.3.3, 6.1), so his theory does not ignore practical influence. What it does not supply is the psychological history and felt manner through which Hume explains belief; Leitgeb explicitly leaves the psychological implementation open (2017, 9--10).

A related temporal distinction matters. Suppositional stability at a time is not, on its own, a complete rule for persistence through an actual learning episode. If \(A\) is initially believed, \(E\) is admitted, and later credence is \(P'=P(\cdot\mid E)\), then \(P'(A)>r\). A later belief in the contrary cannot satisfy the same Humean constraint, since that would require \(P'(W\setminus A)>1/2\). Retaining categorical belief in \(A\) itself, however, also requires appropriate dynamics; merely imposing the static thesis at both times is weaker. This applies Leitgeb's account; it does not reinterpret its basic form.

Belief therefore adds a third requirement beyond updating. A cognitive account must identify the interpreted attitudes that instantiate the stability condition and the processes by which they are acquired. A functional account can address stable use and decision roles without adopting Hume's full phenomenology; a Humean account also owes the relevant manner of conception. Either way, correctly evolving numbers do not by themselves establish the attribution of the attitude. This is the practical point of distinguishing between using an available concept and endorsing a claim that contains it.

\section{Language models: updating performance and three separate diagnoses}

\subsection{What the Bayesian analogy establishes}

Language models make the explanatory gap visible, because a successful updating pattern can be demonstrated in them without any corresponding account of the source of causal understanding. This section examines learning from text and specific published tests; it reports no new experiment. The studies involve different systems and tasks, and the argument about the source channel applies to the report-mediated case specified in section 7.2, not to every system in them. The examples diagnose information sources, available use, and the interpretation of answers. Whether an internal transition is felt is a further question, not an experimental finding drawn from those failures.

Xie et al. (2022) show how an ideal predictor trained on a mixture of hidden Markov models can perform in-context learning as implicit Bayesian inference under specified assumptions of distinguishability and regularity, and synthetic language-model experiments support the proposed correspondence. The result is not a theorem that arbitrary natural-language prompts implement Bayesian inference. Agarwal, Dalal, and Misra (2026, §§2.7, 4, 8) use controlled ``Bayesian wind tunnels'' to compare small Transformers with analytic posterior predictions. They report very close numerical agreement, and their attention analyses and ablations support content-addressable routing. The results cited are from the May 2026 version of their preprint, which was first posted in 2025.

These results warrant two positive claims: a trained sequence model can approximate Bayesian posterior prediction in controlled tasks, and its internal organization can provide a partial counterpart to selective retrieval. It does not follow that gradient training is Bayesian conditionalization over network weights, nor does the predictive agreement establish the experiential acquisition of the task's interpreted content. The mechanism supplies a learned transition; whether it also supplies Hume's impression of that transition is a further question.

\subsection{Experiential grounding and Hume's category of education}

Hume distinguishes custom based on experienced conjunctions from the repetition of ``a mere idea alone'' in education (T 1.3.9.16--19, SBN 115--117). Education can produce assent without any experience of the represented conjunction. His contrast between an artificial cause of assent and its philosophical authority concerns the route by which conviction is produced (T 1.3.9.19; T 1.3.10.1, SBN 117--118); it does not make every report false or every instance of testimony worthless.

Consider the report-mediated case: a model is trained and prompted on a corpus of textual records, with no independent sensor or intervention channel sampling the external events reported. Its evidence about those events comes through reports, not firsthand observation. This follows from the specified acquisition route, not merely from textual encoding, since a sensor stream could itself be serialized as text. Reports can contain measurements, the results of interventions, and reliable causal knowledge. In Hume's human case, education employs ideas whose basic experiential materials may already have been acquired independently; the stipulated route supplies no independent firsthand sampling of the reported events. Education is therefore a useful comparison, though it does not settle whether the model has ideas or internal experience in Hume's sense.

Zečević et al. (2023, 6--7) supply a close contemporary antecedent. Their meta-causal framework distinguishes the represented causal system from the process that produces discourse about it. Their stronger claim, that the success of large language models merely reflects correlations among causal facts, is a conjecture, not a general theorem established by their experiments. My argument adds the separation between evidence about events and the experience that could constitute necessity content, together with the questions of general organization and belief. It does not assume that mediation through language rules out causal knowledge.

Kıcıman et al. (2024, 2--4, 7, 30--32) are a positive interlocutor: they show that language models can supply useful causal domain knowledge and solve a range of causal tasks, including newly constructed cases. They explicitly separate generating causal arguments from genuine causal reasoning, and they discuss construct validity. Their practical proposal combines model output with human checking and formal causal methods. These achievements show why the comparison matters: text-mediated knowledge can support causal work even when the explanation of how the learner acquired causal content remains incomplete. Their findings give no evidence that these authors equate correct answers with Humean impressions.

The evidence channel has a concrete consequence: the frequency with which a conjunction is reported need not equal the frequency with which it occurs. Gordon and Van Durme (2013) document reporting bias in language-based knowledge extraction. Paik et al. (2021, 827--830) find that the object-color distributions of the text-only models they tested track textual frequencies more closely than human color judgments, and that some visually trained representations yield better color distributions under probing. That comparison does not show that adding vision to a particular model creates an impression. McCoy et al. (2024) likewise identify effects of the frequencies of inputs, outputs, and tasks in training. A model may therefore predict reports accurately while misestimating the world they describe. A model of reporting and selection could correct the mismatch, but ordinary updating alone does not identify which distribution its evidence samples.

The role of impressions in constituting content requires a separate argument. Hume recognizes one kind of necessity across the physical and moral domains (T 1.3.14.33, SBN 171). On the historical reading of the source, acquiring its idea does not require a new original impression for every domain of application. It follows that the lack of firsthand impressions of the reported events cannot show that the model lacks the idea of necessity. If the model underwent a relevant felt customary determination in some other domain, perhaps one involving the signs it encounters, that could in principle supply the idea for later application. A transition between word sequences alone establishes neither the requisite feeling nor the acquisition of an idea from it. This is an interpretive consequence of Hume's unified account, not a finding that a language model has such experience.

The two conclusions can now be stated separately. First, the stipulated learner's evidence about reported events is mediated by text. Second, whether its own habitual transitions have ever supplied an impression of reflection remains open. On the reading defended here, the absence of that impression would rule out Hume's source of the idea of necessity, and a successful updating pattern neither establishes nor excludes its occurrence. The point concerns the acquisition history; it is not a demand that the original impression be felt again at every use.

Rival semantic theories give this distinction consequences. Piantadosi and Hill (2022) explain aspects of meaning through relational organization, and Coelho Mollo and Millière (2023, §§5.3--5.5) explain how linguistic mediation can transmit world-involving relations. These proposals address content and reference; establishing either does not by itself establish Hume's felt source. Multimodality can change the evidential channel by adding direct sensory measurements of target events, while leaving the relation between processing and feeling still to be explained. Under a functional reconstruction of Humean feeling, that becomes a question about mechanisms and how they are acquired. Hume's examples of the person blind from birth and the missing shade illustrate his distinction between experiential materials and structural completion (T 1.1.1.9--10, SBN 5--6), but they cannot settle what a modern system experiences. The update rule stays neutral among these accounts.

\subsection{General organization: retained information or failed access?}

The reversal experiments make a more directly testable distinction. Berglund et al. (2024) train models on name--description facts and find marked failures on reversed queries, while noting that reversal succeeds more often when the fact is supplied in context. Correct use in one direction does not ensure use in the other. Proposition 3 shows why a retained state that carries the information can still be deployed unsuccessfully. Proposition 2 applies only if the complete accessible state has actually collapsed a needed distinction, and failure on the reversed question alone cannot establish that stronger premise.

The distinction suggests a focused investigation: compare the learning condition with a supplied-fact condition, determine whether the relevant contrast can be recovered, and intervene on the implicated state or retrieval route. Recovery by prompting can support the distinction between availability and accessibility familiar from Tulving and Pearlstone (1966), though the prompt may also supply new information and must be controlled for. A successful probe alone is still not sufficient evidence of the model's own use. These alternatives make the formal propositions diagnostic, instead of labels that fit any error.

Corr2Cause examines causal reasoning under stated discovery assumptions. Jin et al. (2024, §4.5) report that strong fine-tuned performance deteriorates when variable labels or description templates change. This tests whether use tracks relational structure across irrelevant surface variation, but it neither uniquely identifies the internal respect of resemblance nor shows that every relevant distinction has vanished. Semantically unrelated labels likewise help to separate label meaning from task adaptation in in-context learning (Wei et al. 2023). Such perturbations put pressure on claims of general organization: what the learner can use should be identified at the level of generality claimed.

\subsection{Belief: changed answers and changed commitments}

Sharma et al. (2024, §§3.2--3.3) find that the assistants they tested sometimes abandon correct answers after a user disagrees, or adopt a position the user suggests. Laban et al.'s (2024) FlipFlop experiments similarly show drops in performance after challenges, including a bare request for reassurance. These findings make it unsafe to infer settled endorsement directly from a first answer or a separately elicited confidence report. They do not show that every changed answer is irrational, or that the systems have no beliefs.

Leitgeb's account supplies a sharper conditional test. Suppose that the first answer expresses a belief in \(A\); that the initial belief--credence pair satisfies the Humean constraint; that the challenge represents positive-probability evidence compatible with the whole prior belief set; that revision is conditioning; and that later reports faithfully express a belief satisfying the Humean constraint. Then the later answer cannot express belief in the contrary. Mere compatibility of the challenge with \(A\) is not enough. The experiments do not independently establish the whole belief set, the conditioning event, or the reporting policy. Recomputing a response, changing conversational role, and changing an internal commitment are distinct diagnoses.

Herrmann and Levinstein (2025) propose accuracy, coherence, uniformity, and use as connected standards for attributing beliefs to models. Their discussion of coherence is synchronic and explicitly leaves diachronic stability to further work (§5.5), and they distinguish a belief from its truthful expression. Their approach thus supports a joint investigation of internal representation, use, and behavior, instead of equating a sentence with an attitude. Hume's distinction between a transition within a supposition and belief gives one reason why that evidential task matters.

The three groups of cases establish different things. Reporting bias concerns the relation between sampled reports and the represented world; reversal and renaming concern access and general use; and changes under challenge concern which commitments the answers warrant attributing. None is a behavioral proof that impressions are absent. Together they show why successful Bayesian-like prediction is not sufficient evidence for the coordinated acquisition of a causal judgment.

\section{Why the three conditions must concern the same judgment}

Table 1 records what the strongest counterparts contribute and the acquisition claim that remains distinct from each result. The final column concerns the extended explanatory target; it is not an objection to every model's stated purpose.

\begin{longtable}[]{@{}
  >{\raggedright\arraybackslash}p{(\columnwidth - 4\tabcolsep) * \real{0.1881}}
  >{\raggedright\arraybackslash}p{(\columnwidth - 4\tabcolsep) * \real{0.4059}}
  >{\raggedright\arraybackslash}p{(\columnwidth - 4\tabcolsep) * \real{0.4059}}@{}}
\caption{Three conditions, substantive counterparts, and the remaining explanatory claim.}\tabularnewline
\toprule\noalign{}
\begin{minipage}[b]{\linewidth}\raggedright
Condition
\end{minipage} & \begin{minipage}[b]{\linewidth}\raggedright
What contemporary accounts supply
\end{minipage} & \begin{minipage}[b]{\linewidth}\raggedright
What the result does not establish by itself
\end{minipage} \\
\midrule\noalign{}
\endfirsthead
\toprule\noalign{}
\begin{minipage}[b]{\linewidth}\raggedright
Condition
\end{minipage} & \begin{minipage}[b]{\linewidth}\raggedright
What contemporary accounts supply
\end{minipage} & \begin{minipage}[b]{\linewidth}\raggedright
What the result does not establish by itself
\end{minipage} \\
\midrule\noalign{}
\endhead
\bottomrule\noalign{}
\endlastfoot
Experiential grounding & Evidence-sensitive priors, causal-power assumptions, learned causal theories and intervention roles & Acquisition of Hume's necessity idea from a felt customary transition \\
General organization & Exemplar weighting, sampling, hypothesis construction, and acquired usable resources & Use of the relevant distinction in the further operation attributed to the learner \\
Belief & Stable high credence relative to a whole belief set and qualified decision roles & Psychological realization, experiential history, or identification of a verbal answer with that attitude \\
\end{longtable}

The source condition changes what counts as completing the Humean explanation. Cheng supplies a general causal framework through an innate assumption; GUT learns a conceptual role under a different account of its acquisition. These are different explanatory choices, not failures to calculate a posterior. Section 4 gave a reason to distinguish them: success with perceptual and intervention cues leaves open why the connection is experienced as necessitating. Hume's felt source addresses that question, after his arguments against direct insight into external or voluntary power. The contemporary dispute is whether this experienced aspect constitutes causal content or can be explained separately, and numerical success alone cannot settle that choice.

The source history begins with a minimal discrimination of resembling instances; its later use in an endorsed causal attribution calls for the fuller general organization described in section 2. The other two conditions constrain how the source claim applies. A content attributed on the basis of learning must be the content organized for later use, not a distinction that only the analyst can read. A usable representation must figure in the claimed manner of belief, not solely inside an unendorsed supposition. Conversely, a stable report may concern a different feature from the one the explanation says was learned. Separate successes established for different states or under different external assistance do not automatically combine into one explanation of acquisition.

Consider a learner who at first uses capsule color to predict recovery. Later experience makes the structure of ingredients available; a candidate process recruits that structure for a new treatment; and the resulting judgment governs ordinary planning. The explanation must identify the content directed at the ingredient, establish how it became available, and distinguish its endorsing use from a hypothetical exercise. A supplied ingredient rule, an external decoder, and a stable verbal response each give evidence about part of this history, but none alone establishes the coordinated result. The Humean source question then asks how the causal element of that content was acquired, over and above identifying the ingredient.

This temporal integration is the additional demand the paper places on resource-based accounts of acquisition. It acknowledges the real change in possession of usable resources that these accounts describe, while separating that change from the source of understanding and from belief. A learner can acquire a concept that is used only in fiction, so an acquisition claim need not always include endorsement; but an explanation of a newly acquired causal belief does include it. The target determines which connections are required.

The distinction between normative and descriptive theories remains decisive. Howson's logic and Williamson's evidential norms can succeed without explaining how a child forms concepts, whereas a cognitive claim about the formation of causal judgment incurs the additional requirements. Hume's corrective rules make psychological and normative questions interact, but neither a valid norm nor a successful psychological mechanism automatically answers the other's question. To treat Bayesian updating as a successor to the whole Humean account would conceal this difference in scope.

\section{Conclusion}

Bayesian updating retains a pattern central to Hume's account: evidence changes the strength of expectation. As an updating operation, however, it does not explain how the learner acquires the relevant causal content, organizes it for general use, or holds it in the manner of belief. The strongest contemporary models give substantial accounts of the latter tasks and of some forms of content acquisition, and these additions show why the fuller explanation cannot be read off a posterior.

For the local Humean conception of necessity, the source claim rests on an argument: neither a single observed interaction nor consciousness of willing reveals a necessary productive tie. Learned intervention roles and research on perceptual causality offer serious alternatives, but their success leaves open a question about the experienced force of causal expectation. Hume explains that feature through a felt customary determination arising from discriminated resembling instances, and full general organization articulates its later use in causal attribution. The disagreement concerns the content and acquisition of causal judgment, which remain in dispute even where the numerical predictions agree.

Language models give the distinction a concrete use. For the stipulated text-only learner, evidence about reported events is mediated by reports, and whether a felt transition is present is a separate, open question. Controlled Bayesian-like prediction, reporting bias, asymmetric access, and answer instability therefore call for different diagnoses. The general idea of necessity could in principle be acquired in one domain and used in another, so lacking firsthand evidence about a target event does not by itself rule out possessing it. The paper's critical conclusion is that an explanation of causal judgment must establish the acquisition and coordinated use of the content that turns a successful update into that judgment. Updating alone leaves that work undone.

\appendix

\section{Predictive information and actual deployment}

Xu et al. (2020) formalize the family-relative recoverability distinguished in section 5.2 through predictive \(\mathcal{V}\)-information. A family satisfies optional ignorance when every distribution output by one of its members can also be returned constantly by a member, including without input. For finite outcomes, \(f[X]\) denotes a predictive distribution and \(f[X](Y)\) its probability at the outcome \(Y\). Using base-two logarithms, define

\[H_{\mathcal{V}}(Y\mid X)=\inf_{f\in \mathcal{V}}\ \mathbb{E}_{(X,Y)}\big[-\log_2 f[X](Y)\big]\]

and

\[H_{\mathcal{V}}(Y\mid\varnothing)=\inf_{f\in \mathcal{V}}\ \mathbb{E}_{Y}\big[-\log_2 f[\varnothing](Y)\big].\]

Here \(\varnothing\) is a distinguished no-input symbol. Assuming that the displayed infima are finite, define

\[I_{\mathcal{V}}(X\to Y)=H_{\mathcal{V}}(Y\mid\varnothing)-H_{\mathcal{V}}(Y\mid X).\]

The quantity measures the improvement in optimal expected log loss within the family, not the performance of the deployed controller. With unrestricted predictors it recovers mutual information; computational restrictions can produce different behavior.

In Proposition 3, take an equal mixture of the two parameters and fix \(x=0\). The post-evidence state is \(S=p=\theta\), and the target is \(Y=\theta\), so \(I(S;Y)=1\) bit. For a common predictor family that contains the correct conditional predictor and a no-input predictor returning the uniform Bernoulli distribution, the baseline optimal log loss is one bit and the conditional optimum is zero. Both systems therefore have the same one bit of \(\mathcal{V}\)-information relative to that family, yet their deployed classification errors are zero and one half. The calculation is mine; it is not a result stated in the cited work.

A different family-relative measure could model a different restriction. If B's admissible family contained only constant Bernoulli predictors, its optimal loss would not improve when given \(S\) as input. That additional assumption, however, should be stated; it does not follow merely from B's use of an unsuccessful controller. Nor should a singleton deterministic predictor be used to manufacture the comparison: on the balanced target its log loss can be infinite, which leaves the proposed subtraction undefined. The distinction between logarithmic prediction loss and zero-one classification error also remains in force.

\section{The finite stable-core formulation}

This appendix specializes the finite formulation of section 6 to the conditional threshold \(r=1/2\) (Leitgeb 2014, 139--145). Let \(W\) be a finite nonempty world set with probability measure \(P\). A nonempty core \(K\) is \(P\)-stable when

\[P(K\mid E)>1/2\]

for every \(E\subseteq W\) satisfying \(P(E)>0\) and \(E\cap K\neq\varnothing\). If \(P(K)=1\), require \(K\) to be least under inclusion among probability-one sets. Beliefs are then defined by

\[\mathrm{Bel}(A)\ \Leftrightarrow\ K\subseteq A.\]

The core represents the conjunction of the belief set; it is not uniquely determined by \(P\) alone. These displayed conditions constrain a belief--credence pair at a time. Leitgeb's broader theory also treats belief dynamics and action. Its Humean motivation connects confidence with steadiness instead of identifying a single high marginal probability with the whole of belief (2014, 145n9; 2017, chs. 2, 4, 6).

Define the admitted conditioning events by

\[\Gamma_{\mathrm{Bel}}=\{E\subseteq W: P(E)>0,\ \neg\mathrm{Bel}(W\setminus  E)\}.\]

Since beliefs are the supersets of \(K\), this is exactly the set of positive-probability events that intersect \(K\). In the one-half formulation, the characterization relative to the whole belief set is

\[\mathrm{Bel}(A)\ \Leftrightarrow\ P(A\mid E)>1/2\ \text{ for every } E\in\Gamma_{\mathrm{Bel}}.\]

The equivalence follows directly from the stated core conditions. If \(K\subseteq A\), conditional monotonicity and stability yield the right-hand side. Conversely, if \(K\not\subseteq A\), choose a world in \(K\setminus A\). Every world in the admissible core has positive probability: otherwise stability fails when \(P(K)<1\), while the least-probability-one proviso excludes zero-mass worlds when \(P(K)=1\). Conditioning on that singleton is admitted and gives \(A\) probability zero. This verifies the finite characterization; it is not a new theorem.

An independently specified numerical case shows why the conditioning domain matters. Let \(P(w_1,w_2,w_3)=(.98,.01,.01)\), \(K=\{w_1\}\), and \(A=\{w_1,w_2\}\). Then \(A\) is believed because \(K\subseteq A\). Although conditioning on \(E_0=\{w_2,w_3\}\) reduces \(P(A)\) from \(.99\) to \(.5\), \(E_0\) is disjoint from \(K\); its negation is believed, so it is excluded from \(\Gamma_{\mathrm{Bel}}\). Across the admitted events, \(P(A\mid E)\) is at least \(98/99\). The proposition therefore satisfies the Humean requirement while failing the stronger, proposition-relative \(P\)-stability test, which admits every positive-probability condition that intersects \(A\) itself.

The example is not a counterexample to the Humean thesis. It shows that marginal probability, stability relative to one proposition, and stability relative to a whole belief set are different conditions. The stable core specifies which suppositions remain admissible. Adopting this bridge yields substantive mathematical constraints on categorical commitment, but demonstrating that a particular system realizes the relevant doxastic organization requires a further interpretive argument. That is a division of explanatory work, not a rejection of a formal account of belief.

\section*{References}
\begingroup\small\sloppy\setlength{\parskip}{4pt}
\bibentry Agarwal, Naman, Siddhartha R. Dalal, and Vishal Misra. 2026. ``The Bayesian Geometry of Transformer Attention.'' Preprint, arXiv:2512.22471v5, May 16; first submitted December 27, 2025. \url{https://doi.org/10.48550/arXiv.2512.22471}.

\bibentry Belinkov, Yonatan. 2022. ``Probing Classifiers: Promises, Shortcomings, and Advances.'' \emph{Computational Linguistics} 48 (1): 207--219. \url{https://doi.org/10.1162/coli_a_00422}.

\bibentry Berglund, Lukas, Meg Tong, Max Kaufmann, Mikita Balesni, Asa Cooper Stickland, Tomasz Korbak, and Owain Evans. 2024. ``The Reversal Curse: LLMs Trained on `A Is B' Fail to Learn `B Is A'.'' \emph{International Conference on Learning Representations}. \url{https://proceedings.iclr.cc/paper_files/paper/2024/file/5178b2f2d7c44aa390c0777dc77b3f0c-Paper-Conference.pdf}.

\bibentry Broughton, Janet. 2000. ``Explaining General Ideas.'' \emph{Hume Studies} 26 (2): 279--289. \url{https://doi.org/10.1353/hms.2000.a385726}.

\bibentry Buckner, Cameron J. 2024. \emph{From Deep Learning to Rational Machines: What the History of Philosophy Can Teach Us about the Future of Artificial Intelligence}. Oxford: Oxford University Press. \url{https://doi.org/10.1093/oso/9780197653302.001.0001}.

\bibentry Carey, Susan. 2009. \emph{The Origin of Concepts}. Oxford: Oxford University Press. \url{https://doi.org/10.1093/acprof:oso/9780195367638.001.0001}.

\bibentry Cheng, Patricia W. 1997. ``From Covariation to Causation: A Causal Power Theory.'' \emph{Psychological Review} 104 (2): 367--405. \url{https://doi.org/10.1037/0033-295X.104.2.367}.

\bibentry Chollet, François. 2019. ``On the Measure of Intelligence.'' Preprint, arXiv:1911.01547. \url{https://doi.org/10.48550/arXiv.1911.01547}.

\bibentry Coelho Mollo, Dimitri, and Raphaël Millière. 2023. ``The Vector Grounding Problem.'' Preprint, arXiv:2304.01481v1, April 4. \url{https://arxiv.org/abs/2304.01481v1}.

\bibentry Dasgupta, Ishita, Eric Schulz, and Samuel J. Gershman. 2017. ``Where Do Hypotheses Come From?'' \emph{Cognitive Psychology} 96: 1--25. \url{https://doi.org/10.1016/j.cogpsych.2017.05.001}.

\bibentry Earman, John. 1992. \emph{Bayes or Bust? A Critical Examination of Bayesian Confirmation Theory}. Cambridge, MA: MIT Press.

\bibentry Fodor, Jerry A. 2008. \emph{LOT 2: The Language of Thought Revisited}. Oxford: Oxford University Press. \url{https://doi.org/10.1093/acprof:oso/9780199548774.001.0001}.

\bibentry Garrett, Don. 1997. \emph{Cognition and Commitment in Hume's Philosophy}. New York: Oxford University Press. \url{https://doi.org/10.1093/oso/9780195097214.001.0001}.

\bibentry Goodman, Noah D., Tomer D. Ullman, and Joshua B. Tenenbaum. 2011. ``Learning a Theory of Causality.'' \emph{Psychological Review} 118 (1): 110--119. \url{https://doi.org/10.1037/a0021336}.

\bibentry Gordon, Jonathan, and Benjamin Van Durme. 2013. ``Reporting Bias and Knowledge Acquisition.'' In \emph{Proceedings of the 2013 Workshop on Automated Knowledge Base Construction}, 25--30. \url{https://doi.org/10.1145/2509558.2509563}.

\bibentry Griffiths, Thomas L., and Joshua B. Tenenbaum. 2009. ``Theory-Based Causal Induction.'' \emph{Psychological Review} 116 (4): 661--716. \url{https://doi.org/10.1037/a0017201}.

\bibentry Harding, Jacqueline. 2023. ``Operationalising Representation in Natural Language Processing.'' \emph{The British Journal for the Philosophy of Science}, advance online publication. \url{https://doi.org/10.1086/728685}.

\bibentry Harnad, Stevan. 1990. ``The Symbol Grounding Problem.'' \emph{Physica D: Nonlinear Phenomena} 42 (1--3): 335--346. \url{https://doi.org/10.1016/0167-2789(90)90087-6}.

\bibentry Herrmann, Daniel A., and Benjamin A. Levinstein. 2025. ``Standards for Belief Representations in LLMs.'' \emph{Minds and Machines} 35: 5. \url{https://doi.org/10.1007/s11023-024-09709-6}.

\bibentry Hewitt, John, and Percy Liang. 2019. ``Designing and Interpreting Probes with Control Tasks.'' In \emph{Proceedings of EMNLP-IJCNLP}, 2733--2743. \url{https://doi.org/10.18653/v1/D19-1275}.

\bibentry Howson, Colin. 2000. \emph{Hume's Problem: Induction and the Justification of Belief}. Oxford: Clarendon Press. \url{https://doi.org/10.1093/0198250371.001.0001}.

\bibentry Hume, David. 1739--40/1978. \emph{A Treatise of Human Nature}. Edited by L. A. Selby-Bigge. Second edition revised by P. H. Nidditch. Oxford: Clarendon Press. SBN refers to this edition's pagination; paragraph numbering follows the 2007 critical edition. Text consulted at \url{https://davidhume.org/texts/t/}.

\bibentry Hume, David. 1748/1975. \emph{An Enquiry concerning Human Understanding}. In \emph{Enquiries concerning Human Understanding and concerning the Principles of Morals}. Edited by L. A. Selby-Bigge. Third edition revised by P. H. Nidditch. Oxford: Clarendon Press. SBN refers to this edition's pagination; paragraph numbering follows the 2000 critical edition. Text consulted at \url{https://davidhume.org/texts/e/}.

\bibentry Hume, David. 2000. \emph{An Enquiry concerning Human Understanding: A Critical Edition}. Edited by Tom L. Beauchamp. Oxford: Clarendon Press.

\bibentry Hume, David. 2007. \emph{A Treatise of Human Nature: A Critical Edition}. Edited by David Fate Norton and Mary J. Norton. Vol. 1, \emph{Texts}. Oxford: Clarendon Press.

\bibentry Jeffrey, Richard C. 1990. \emph{The Logic of Decision}. 2nd ed. Chicago: University of Chicago Press.

\bibentry Jin, Zhijing, Jiarui Liu, Zhiheng Lyu, Spencer Poff, Mrinmaya Sachan, Rada Mihalcea, Mona Diab, and Bernhard Schölkopf. 2024. ``Can Large Language Models Infer Causation from Correlation?'' \emph{International Conference on Learning Representations}. \url{https://proceedings.iclr.cc/paper_files/paper/2024/file/7b75a7339dfb256ee4b4bec028a6890b-Paper-Conference.pdf}.

\bibentry Jones, Matt, and Bradley C. Love. 2011. ``Bayesian Fundamentalism or Enlightenment? On the Explanatory Status and Theoretical Contributions of Bayesian Models of Cognition.'' \emph{Behavioral and Brain Sciences} 34 (4): 169--188. \url{https://doi.org/10.1017/S0140525X10003134}.

\bibentry Joyce, James M. 1998. ``A Nonpragmatic Vindication of Probabilism.'' \emph{Philosophy of Science} 65 (4): 575--603. \url{https://doi.org/10.1086/392661}.

\bibentry Kemp, Charles, A. Perfors, and Joshua B. Tenenbaum. 2007. ``Learning Overhypotheses with Hierarchical Bayesian Models.'' \emph{Developmental Science} 10 (3): 307--321. \url{https://doi.org/10.1111/j.1467-7687.2007.00585.x}.

\bibentry Kıcıman, Emre, Robert Osazuwa Ness, Amit Sharma, and Chenhao Tan. 2024. ``Causal Reasoning and Large Language Models: Opening a New Frontier for Causality.'' Preprint, arXiv:2305.00050v3, August 20; first submitted April 28, 2023. \url{https://doi.org/10.48550/arXiv.2305.00050}.

\bibentry Kominsky, Jonathan F., and Susan Carey. 2024. ``Infants' Representations of Michottean Triggering Events.'' \emph{Cognition} 250: 105844. \url{https://doi.org/10.1016/j.cognition.2024.105844}.

\bibentry Laban, Philippe, Lidiya Murakhovs'ka, Caiming Xiong, and Chien-Sheng Wu. 2024. ``Are You Sure? Challenging LLMs Leads to Performance Drops in the FlipFlop Experiment.'' Preprint, arXiv:2311.08596v2, February 21; first submitted November 14, 2023. \url{https://arxiv.org/abs/2311.08596v2}.

\bibentry Landes, Jürgen, and Jon Williamson. 2013. ``Objective Bayesianism and the Maximum Entropy Principle.'' \emph{Entropy} 15 (9): 3528--3591. \url{https://doi.org/10.3390/e15093528}.

\bibentry Leitgeb, Hannes. 2014. ``The Stability Theory of Belief.'' \emph{Philosophical Review} 123 (2): 131--171. \url{https://doi.org/10.1215/00318108-2400575}.

\bibentry Leitgeb, Hannes. 2015. ``I---The Humean Thesis on Belief.'' \emph{Aristotelian Society Supplementary Volume} 89 (1): 143--185. \url{https://doi.org/10.1111/j.1467-8349.2015.00248.x}.

\bibentry Leitgeb, Hannes. 2017. \emph{The Stability of Belief: How Rational Belief Coheres with Probability}. Oxford: Oxford University Press. \url{https://doi.org/10.1093/acprof:oso/9780198732631.001.0001}.

\bibentry Leslie, Alan M., and Stephanie Keeble. 1987. ``Do Six-Month-Old Infants Perceive Causality?'' \emph{Cognition} 25 (3): 265--288. \url{https://doi.org/10.1016/S0010-0277(87)80006-9}.

\bibentry Lu, Hongjing, Alan L. Yuille, Mimi Liljeholm, Patricia W. Cheng, and Keith J. Holyoak. 2008. ``Bayesian Generic Priors for Causal Learning.'' \emph{Psychological Review} 115 (4): 955--984. \url{https://doi.org/10.1037/a0013256}.

\bibentry Margolis, Eric, and Stephen Laurence. 2011. ``Learning Matters: The Role of Learning in Concept Acquisition.'' \emph{Mind \& Language} 26 (5): 507--539. \url{https://doi.org/10.1111/j.1468-0017.2011.01429.x}.

\bibentry McCoy, R. Thomas, Shunyu Yao, Dan Friedman, Mathew D. Hardy, and Thomas L. Griffiths. 2024. ``Embers of Autoregression Show How Large Language Models Are Shaped by the Problem They Are Trained to Solve.'' \emph{Proceedings of the National Academy of Sciences} 121 (41): e2322420121. \url{https://doi.org/10.1073/pnas.2322420121}.

\bibentry Paik, Cory, Stéphane Aroca-Ouellette, Alessandro Roncone, and Katharina Kann. 2021. ``The World of an Octopus: How Reporting Bias Influences a Language Model's Perception of Color.'' In \emph{Proceedings of EMNLP}, 823--835. \url{https://doi.org/10.18653/v1/2021.emnlp-main.63}.

\bibentry Perfors, A. 2012. ``Bayesian Models of Cognition: What's Built in After All?'' \emph{Philosophy Compass} 7 (2): 127--138. \url{https://doi.org/10.1111/j.1747-9991.2011.00467.x}.

\bibentry Piantadosi, Steven T. 2021. ``The Computational Origin of Representation.'' \emph{Minds and Machines} 31: 1--58. \url{https://doi.org/10.1007/s11023-020-09540-9}.

\bibentry Piantadosi, Steven T., and Felix Hill. 2022. ``Meaning without Reference in Large Language Models.'' Preprint, arXiv:2208.02957v2, August 12. \url{https://arxiv.org/abs/2208.02957v2}.

\bibentry Piantadosi, Steven T., Joshua B. Tenenbaum, and Noah D. Goodman. 2012. ``Bootstrapping in a Language of Thought: A Formal Model of Numerical Concept Learning.'' \emph{Cognition} 123 (2): 199--217. \url{https://doi.org/10.1016/j.cognition.2011.11.005}.

\bibentry Rips, Lance J., Jennifer Asmuth, and Amber Bloomfield. 2013. ``Can Statistical Learning Bootstrap the Integers?'' \emph{Cognition} 128 (3): 320--330. \url{https://doi.org/10.1016/j.cognition.2013.04.001}.

\bibentry Ritchie, J. Brendan, David Michael Kaplan, and Colin Klein. 2019. ``Decoding the Brain: Neural Representation and the Limits of Multivariate Pattern Analysis in Cognitive Neuroscience.'' \emph{The British Journal for the Philosophy of Science} 70 (2): 581--607. \url{https://doi.org/10.1093/bjps/axx023}.

\bibentry Rolfs, Martin, Michael Dambacher, and Patrick Cavanagh. 2013. ``Visual Adaptation of the Perception of Causality.'' \emph{Current Biology} 23 (3): 250--254. \url{https://doi.org/10.1016/j.cub.2012.12.017}.

\bibentry Rule, Joshua S., and Steven T. Piantadosi. 2025. ``The End of Radical Concept Nativism.'' Preprint, arXiv:2505.18277v2, revised July 9. \url{https://doi.org/10.48550/arXiv.2505.18277}.

\bibentry Sanborn, Adam N., and Nick Chater. 2016. ``Bayesian Brains without Probabilities.'' \emph{Trends in Cognitive Sciences} 20 (12): 883--893. \url{https://doi.org/10.1016/j.tics.2016.10.003}.

\bibentry Sharma, Mrinank, Meg Tong, Tomasz Korbak, David Duvenaud, Amanda Askell, Samuel R. Bowman, Newton Cheng, Esin Durmus, Zac Hatfield-Dodds, Scott R. Johnston, Shauna Kravec, Timothy Maxwell, Sam McCandlish, Kamal Ndousse, Oliver Rausch, Nicholas Schiefer, Da Yan, Miranda Zhang, and Ethan Perez. 2024. ``Towards Understanding Sycophancy in Language Models.'' \emph{International Conference on Learning Representations}. \url{https://proceedings.iclr.cc/paper_files/paper/2024/file/0105f7972202c1d4fb817da9f21a9663-Paper-Conference.pdf}.

\bibentry Shea, Nicholas. 2018. \emph{Representation in Cognitive Science}. Oxford: Oxford University Press. \url{https://doi.org/10.1093/oso/9780198812883.001.0001}.

\bibentry Shi, Lei, Thomas L. Griffiths, Naomi H. Feldman, and Adam N. Sanborn. 2010. ``Exemplar Models as a Mechanism for Performing Bayesian Inference.'' \emph{Psychonomic Bulletin \& Review} 17 (4): 443--464. \url{https://doi.org/10.3758/PBR.17.4.443}.

\bibentry Stigler, Stephen M. 2013. ``The True Title of Bayes's Essay.'' \emph{Statistical Science} 28 (3): 283--288. \url{https://doi.org/10.1214/13-STS438}.

\bibentry Tulving, Endel, and Zena Pearlstone. 1966. ``Availability versus Accessibility of Information in Memory for Words.'' \emph{Journal of Verbal Learning and Verbal Behavior} 5 (4): 381--391. \url{https://doi.org/10.1016/S0022-5371(66)80048-8}.

\bibentry Wei, Jerry, Jason Wei, Yi Tay, Dustin Tran, Albert Webson, Yifeng Lu, Xinyun Chen, Hanxiao Liu, Da Huang, Denny Zhou, and Tengyu Ma. 2023. ``Larger Language Models Do In-context Learning Differently.'' Preprint, arXiv:2303.03846v2, March 8. \url{https://arxiv.org/abs/2303.03846v2}.

\bibentry Wenmackers, Sylvia, and Jan-Willem Romeijn. 2016. ``New Theory about Old Evidence: A Framework for Open-Minded Bayesianism.'' \emph{Synthese} 193 (4): 1225--1250. \url{https://doi.org/10.1007/s11229-014-0632-x}.

\bibentry Williamson, Jon. 2010. \emph{In Defence of Objective Bayesianism}. Oxford: Oxford University Press. \url{https://doi.org/10.1093/acprof:oso/9780199228003.001.0001}.

\bibentry Wu, Yiling. 2026. ``Hume's Two Definitions and the Representational Conditions of Causal Judgment.'' Unpublished manuscript, BridgeM, Inc.

\bibentry Xie, Sang Michael, Aditi Raghunathan, Percy Liang, and Tengyu Ma. 2022. ``An Explanation of In-context Learning as Implicit Bayesian Inference.'' \emph{International Conference on Learning Representations}. Version consulted: arXiv:2111.02080v6. \url{https://doi.org/10.48550/arXiv.2111.02080}.

\bibentry Xu, Yilun, Shengjia Zhao, Jiaming Song, Russell Stewart, and Stefano Ermon. 2020. ``A Theory of Usable Information under Computational Constraints.'' \emph{International Conference on Learning Representations}. \url{https://doi.org/10.48550/arXiv.2002.10689}.

\bibentry Zečević, Matej, Moritz Willig, Devendra Singh Dhami, and Kristian Kersting. 2023. ``Causal Parrots: Large Language Models May Talk Causality But Are Not Causal.'' \emph{Transactions on Machine Learning Research}. Published-format version, arXiv:2308.13067v1. \url{https://arxiv.org/abs/2308.13067v1}.
\par\endgroup
\end{document}